\documentclass[a4paper]{article}

\usepackage{INTERSPEECH2022}
\usepackage[scientific-notation=true]{siunitx}

\usepackage{xcolor}

\usepackage{hyperref}
\usepackage{multirow}
\usepackage{url}
\usepackage{breakurl}

\title{A Study of Gender Impact in Self-supervised Models \\
for Speech-to-Text Systems}

\name{Marcely Zanon Boito$^1$, Laurent Besacier$^2$, Natalia Tomashenko$^1$, Yannick Estève$^1$}
\address{
  $^1$Laboratoire d'Informatique d'Avignon (LIA) - Avignon University, Avignon - France\\
  $^2$NAVER LABS Europe, Meylan - France}
\email{\{name.last-name\}@{\{univ-avignon.fr$^1$, naverlabs.com$^2$\}}
}

\begin{document}

\maketitle
\begin{abstract}
Self-supervised models for speech processing emerged recently as popular foundation blocks in speech processing pipelines. These models are pre-trained on unlabeled audio data and then used in speech processing downstream tasks such as automatic speech recognition~(ASR) or speech translation~(ST). Since these models are now used in research and industrial systems alike, it becomes necessary to understand the impact caused by some features such as gender distribution within pre-training data. Using French as our investigation language, we train and compare \textit{gender-specific} wav2vec~2.0 models against models containing different degrees of gender balance in their pre-training data. The comparison is performed by applying these models to two speech-to-text downstream tasks:~ASR and ST. Results show the type of downstream integration matters. We observe lower overall performance using \textit{gender-specific} pre-training before fine-tuning an end-to-end ASR system. 
However, when self-supervised models are used as feature extractors, the overall ASR and ST results follow more complex patterns in which the balanced pre-trained model does not necessarily lead to the best results.
Lastly, our crude `fairness' metric, the relative performance difference measured between female and male test sets, does not display a strong variation from balanced to gender-specific pre-trained wav2vec~2.0 models.
\end{abstract}

\section{Introduction}

Recently, models based on self-supervised learning~(SSL) for speech processing~\cite{schneider2019wav2vec, hsu2021hubert, baevski2019effectiveness, baevski2020wav2vec} emerged as popular foundation blocks in speech 
pipelines. These models are large trainable networks with millions or even billions~\cite{babu2021xls} of parameters that are trained on unlabeled audio data, hence \textit{self-supervised}. 
The goal of training these models is providing a powerful and reusable abstraction block, able to process raw audio in a given language or in multilingual settings~\cite{conneau2020unsupervised,babu2021xls}, producing a richer audio representation for the downstream tasks to train with, compared to standard features such as MFCCs or filterbanks. Recent work found considerable performance gains and/or state-of-the-art performance by including these blocks in downstream tasks. Most of them focused in automatic speech recognition~(ASR)~\cite{kawakami-etal-2020-learning,schneider2019wav2vec, hsu2021hubert, baevski2019effectiveness, baevski2020wav2vec}, but recent speech benchmarks~\cite{evain21_interspeech,evain2021task,yang21c_interspeech} cover tasks such as speech translation~(ST), spoken language understanding, emotion recognition from speech and more. 
Regarding the use of the self-supervised block in downstream tasks, they can be used either as:
(1)~a feature extractor, with no fine-tuning of the trained weights during downstream task training being performed; or as (2)~a speech encoder, with fine-tuning of the entire model in an end-to-end fashion, together with the additional task-specific modules.

However, independently of the approach used for fine-tuning, one can expect that the characteristics of the speech data used for pre-training may influence the performance of the 
downstream task models.
In this work, we focus on possible gender bias introduced by unbalanced speech data used to pre-train SSL models. 
We train \textit{gender-specific} wav2vec~2.0~\cite{baevski2020wav2vec} models for the French language, and we apply them, together with three off-the-shelf 
wav2vec~2.0 models with different degrees of gender balance, to two downstream tasks:~ASR and ST. For the downstream tasks training, we use the mTEDx dataset~\cite{salesky21_interspeech}, whose gender annotation for the French subset is also a contribution of this work. 
Moreover, 
we explore the aforementioned strategies (1) and (2) for ASR, and (1) for ST, aiming to also investigate their impact in the gender-specific performance of the task models. Our results show that the type of downstream integration matters.
We observe lower overall performance using gender-specific pre-training before an ASR system based on strategy~(1). However, when SSL models are used as 
feature extractors~(2), the overall ASR and ST results 
follow more complex patterns.

Gender bias in speech-to-text systems is defined as a systematic worse recognition for a gender category~\cite{feng2021quantifying}. 
Pioneer work for ASR~\cite{adda2005speech} 
found better performance on women's voices, while a preliminary research on YouTube automatic caption system found better recognition rate for male speech~\cite{tatman2017gender}, and no gender difference in a follow-up study~\cite{tatman17-dialect-gender-effect}. Recent work on hybrid ASR systems observed that gender imbalance in data could lead to decreased ASR performance on the gender category least represented~\cite{Garnerin:2019:GRF:3347449.3357480}, but a posterior work from the same authors observed that ASR trained on audio-books is rather robust to gender imbalance~\cite{garnerin-etal-2021-investigating}, and that other factors~(such as random seed and individuals in the training data) have an important impact as well. Methodological work discussing how to measure fairness in ASR~\cite{DBLP:journals/corr/abs-2109-09061}, and position papers on biases in ASR~\cite{https://doi.org/10.48550/arxiv.2202.12603} were also published recently. Regarding gender bias in ST systems, recent work focused on the content of the generated text, rather than speech itself~\cite{savoldi2022under, costa2020evaluating}.

To our knowledge, the only other investigation of gender bias in 
models for speech processing is the work of Meng et al.~\cite{dontspeaktofast}, but they did not experiment with wav2vec~2.0 SSL models, and did not consider ST and ASR tasks, evaluating downstram performance only on phoneme recognition. In addition, they did not compare strategies~(1) and~(2) mentioned earlier, and their SSL models were trained only on small subsets of Librispeech~(100h), whereas we investigate with models trained on much more data. Lastly, we 
acknowledge that the definition of gender as a binary category is somehow reducing, but we find ourselves limited by the data and metadata available.

This paper is organized as follows. Section~\ref{sec:data} presents the data used for pre-training and downstream tasks, 
and Section~\ref{sec:ssl} describes the SSL models. Section~\ref{sec:asr} and \ref{sec:st} present respectively our ASR and ST results. Section~\ref{sec:discussion} summarizes our findings. 
\section{Data}\label{sec:data}

\noindent \textbf{Pre-training Data.}
For building gender-specific datasets for SSL training, we use the same data from the \textit{LeBenchmark}~\cite{evain21_interspeech,evain2021task}. They gathered a massive amount of French audio of different speech styles, and with rich metadata information.\footnote{Data available at \url{https://github.com/LeBenchmark/NeurIPS2021/tree/main/data_preprocessing}} We select all ten datasets that had gender information, which 
resulted in 1,041\,h of female speech, and 1,006\,h of male speech after down-sampling the EPAC dataset for keeping 
the total duration equivalent between both sets. 
Table~\ref{tab:pretrainingdata} presents key statistics.

\begin{table}
    \centering
    \caption{Statistics for the male/female 
    datasets used for SSL training on French speech. Duration written as hours:minutes.}
    \resizebox{\columnwidth}{!}{
    \begin{tabular}{lccc}
    \toprule
\textbf{Dataset} & \textbf{Duration}                                      & \textbf{\# speakers} & \textbf{Speech Style}                                                \\\midrule
\textbf{MLS}~\cite{pratap_mls_2020}              & 520:13 / 576:29                                   & 80 / 98              & Read                                                                \\
\textbf{Att\_Hack}~\cite{le_moine_att-hack_2020}        & 12:07 / 14:54                                    & 9 / 11               & \begin{tabular}[c]{@{}c@{}}Acted / \\ Emotional\end{tabular}        \\
\textbf{CaFE}~\cite{gournay_canadian_2018}             & 00:32 / 00:36                                    & 6 / 6                & \begin{tabular}[c]{@{}c@{}}Acted / \\ Emotional\end{tabular}        \\
\textbf{CFPP2000}~\cite{branca-rosoff_discours_2012}         & 00:11 / 01:41                                    & 2 / 4                & Spontaneous                                                         \\
\textbf{ESLO2}~\cite{old_eshkol-taravella_grand_2011}            & 17:06 / 16:57                                   & 68 / 120             & Spontaneous                                                         \\
\textbf{EPAC}~\cite{old_esteve_epac_2010}             & 413:41 / 385:52                                  & Unknown              & \begin{tabular}[c]{@{}c@{}}Radio \\ Broadcasts\end{tabular}         \\
\textbf{GEMEP}~\cite{old_banziger_introducing_2012}            & 00:24 / 00:26                                    & 5 / 5                & \begin{tabular}[c]{@{}c@{}}Acted / \\ Emotional\end{tabular}        \\
\textbf{PORTMEDIA}~\cite{old_lefevre_robustesse_2012}        & 19:08 / 19:50                                   & 84 / 109             & \begin{tabular}[c]{@{}c@{}}Acted telephone \\ dialogue\end{tabular} \\
\textbf{TCOF}~\cite{TCOF_ortolang}             & 10:47 / 11:22                                   & 117 / 162            & Spontaneous                                                         \\
\textbf{NCCFr}~\cite{torreira_nijmegen_2010}            & 12:44 / 12:59                                   & 24 / 21              & Spontaneous                                                         \\\midrule
\textbf{TOTAL (M/F)}   & \multicolumn{1}{l}{\textbf{1,006:59 / 1,041:11}} & -                    & -                                                                  \\\bottomrule
\end{tabular}}
    \label{tab:pretrainingdata}
\end{table}

\noindent \textbf{Speech-to-text Data.}
For the speech-to-text downstream tasks, we use the mTEDx dataset~\cite{salesky21_interspeech}. Since there was no gender information available, we manually annotated the \textit{fr-fr} corpus by checking the speaker names, and by watching some of the videos online. Thus, one contribution of this work is the gender annotation in the mTEDx \textit{fr-*} corpus that is now included in its latest release.\footnote{Available at \url{http://www.openslr.org/100}}
For ASR, we down-sample the \textit{fr-fr} subset~(172\,h), creating a gender balanced subset: we sampled the data by gender, reaching roughly 38\,h of gender-specific speech in the training set, which corresponds to a half of the total amount of female speech in the original content. We use only a half of this number because we also created 68\,h gender-specific ASR subsets that we intend to compare against this one in future work focusing on gender-bias in ASR fine-tuning. For the validation set of this balanced subset, the male speech was up-sampled using the unused male entries from the original training set. The test set was kept the same.
For ST, we use the English, Portuguese and Spanish subsets~(respectively \textit{fr-\{en,pt,es\}}; 48\,h, 35\,h, 23\,h).\footnote{The original paper~\cite{salesky21_interspeech} reports respectively 50\,h, 38\,h, 25\,h, but we compute statistics on speech segments only (not full audio duration).}
We highlight that the data for ST is a subset of the \textit{fr-fr}: the validation and test sets are the same. Table~\ref{tab:asrdata} presents the statistics.

\begin{table}
\centering
\caption{Statistics for the \textit{fr-fr} mTEDx with gender annotation (M=male;F=female;B=speakers of both genders present), its 
balanced version~(ASR), and the three ST subsets. Duration written as hours:minutes.}
\scalebox{0.80}{
\begin{tabular}{l|lccc|c}\toprule
\multicolumn{6}{c}{\textbf{Original Content (fr-fr)}}                              \\
                       & \multicolumn{1}{c}{} & \textbf{M}         & \textbf{F}        & \textbf{B}    & \textbf{All}       \\\midrule
\multirow{2}{*}{train} & \# speakers          & 550       & 388      & 4       & 942       \\
                       & Duration             & 100:02 & 68:28 & 0:44 & 169:14 \\
\multirow{2}{*}{valid} & \# speakers          & 5         & 7        & -       & 12        \\
                       & Duration             & 0:38   & 1:00  & -       & 1:38   \\
\multirow{2}{*}{test}  & \# speakers          & 6         & 4        & -       & 10        \\
                       & Duration             & 0:54   & 0:39  & -       & 1:33   \\\midrule
\multicolumn{6}{c}{\textbf{Balanced Dataset (ASR) }}                              \\
\midrule
\multirow{2}{*}{train} & \# speakers          & 550       & 388      & -       & 938       \\
                       & Duration             & 34:09  & 34:09 & -       & 68:17  \\
\multirow{2}{*}{valid} & \# speakers          & 11        & 7        & -       & 18        \\
                       & Duration             & 0:30   & 0:30  & -       & 1:00  \\\midrule
\multicolumn{6}{c}{\textbf{Translation Datasets (ST) }}                              \\
\midrule
\multirow{2}{*}{fr-en (train)} & \# speakers & 146        & 102        & 2             & 250          \\
                                        & Duration    & 26:28   & 18:14   & 0:22       & 45:04     \\
\multirow{2}{*}{fr-es (train)} & \# speakers & 110        & 86         & -             & 196          \\
                                        & Duration    & 17:59   & 14:31   & -             & 32:30     \\
\multirow{2}{*}{fr-pt (train)} & \# speakers & 57         & 55         & -             & 112          \\
                                        & Duration    & 10:39   & 9:22    & -             & 20:01     \\\midrule
\end{tabular}}
\label{tab:asrdata}
\end{table}

\section{Self-Supervised Learning Models}\label{sec:ssl}

We train two gender-specific wav2vec~2.0 \textit{large} models using the 1K datasets presented in Section~\ref{sec:data}, and using the same 
hyperparameters from the original wav2vec~2.0~\cite{baevski2020wav2vec}. We train them using the \textit{fairseq} library~\cite{ott2019fairseq}, and for $125$K updates on $16$ Nvidia Tesla V100~(32GB).\footnote{Due to training instability in the fairseq library, we were unable to reach $200$K updates on the gender-specific models. However, we observed that the trained models at $125$K updates achieve a loss that is lower than the one achieved by the LB-1K-Large model on the same validation set. We thus believe that these models are comparable.} These gender-specific models are added to the collection of pre-trained wav2vec~2.0 models for the French language from the \textit{LeBenchmark}~(LB)~\cite{evain21_interspeech,evain2021task}, and they are available for download at \textit{HuggingFace}.\footnote{\url{https://huggingface.co/LeBenchmark}} 
In this work, we investigate the impact of gender distribution in SSL 
models' pre-training data, focusing on speech-to-text downstream tasks. 
We compare the gender-specific models described above against three models of equal capacity from the 
LB collection~(1K-Large, 3K-Large and 7K-Large). These models are relevant 
because they present different degrees of gender balance in their pre-training data.
A summary of all models 
is presented in Table~\ref{tab:w2v2modelsdescription}.

\begin{table}
\centering
\caption{List of wav2vec~2.0 models, number of updates and hours used for pre-training. The last three columns present the percentage of male~(M), female~(F) and unknown gender~(U) speech present in the pre-training dataset.}
\resizebox{\columnwidth}{!}{
\begin{tabular}{lcc|ccc}
\toprule
\multicolumn{1}{c}{\textbf{Model}} & \textbf{\# updates} & \multicolumn{1}{c|}{\textbf{\# hours}} & \textbf{M,\%} & \textbf{F,\%} & \textbf{U,\%} \\\midrule
F-1K-Large                              & 125K                & 1,041                             &  0               & 100             & 0               \\
M-1K-Large                              & 125K                & 1,006                                 & 100             & 0               & 0               \\
LB-1K-Large                              & 200K                & 1,096                                & 47.4           & 52.5           & 0              \\
LB-3K-Large                                & 500K                & 2,933                                & 62.2           & 35.2           & 2.5            \\
LB-7K-Large                                & 500K                & 7,739                              & 23.9           & 13.4           & 62.6         \\\bottomrule 
\end{tabular}
}
\label{tab:w2v2modelsdescription}
\end{table}

\section{Automatic Speech Recognition}\label{sec:asr}

We experiment with two different ASR models:  a hybrid deep neural network~(DNN) hidden markov model~(HMM), and an end-to-end model.
For DNN-HMM models, the SSL block is used as a feature extractor, and for end-to-end  models it is used as a trainable speech encoder.
The performance is evaluated in terms of word error rate (WER).
The relative difference of WERs between female and male datasets is computed  as Equation~\ref{eq:fairness}, and it can be understood as a basic fairness metric.
\begin{equation}\label{eq:fairness}
\Delta_{rel}=100\frac{WER_\text{female}-WER_\text{male}}{0.5 \times (WER_\text{female}+WER_\text{male})}
\end{equation}

\subsection{Hybrid ASR}

We trained five hybrid DNN-HMM acoustic models using features extracted by the SSL models described in Section~\ref{sec:ssl}.
All models were trained on the balanced dataset~(68\,h) using the Kaldi toolkit~\cite{povey2011kaldi} with a  factorized time delay neural network~(TDNN-F) architecture~\cite{povey2018semi,peddinti2015time}.
The models have 12 TDNN-F layers~(1,024-dimensional, with projection dimension of 128) and a 3K dimensional output layer. 
They were trained using lattice-free maximum mutual information~(LF-MMI)~\cite{povey2016purely}  and cross-entropy criterion.
Speed and volume perturbations have been applied for data augmentation, and 100-dimensional speaker i-vectors were appended to the input features. Finally, a trigram language model~(LM) 
with a 82K vocabulary was used.

Results are presented in the top portion~(a) of Table~\ref{tab:resultsasr}.
We observe that models trained on features extracted from gender-specific pre-trained models performed very closely to the one using features from a model with balanced pre-training~(LB-1K-Large model).
Following intuition, we also observe that among the SSL models trained on 1K hours, the best results for each gender-specific dataset~(M and F columns) were obtained when the gender of the SSL model matched the gender of the speakers in the dataset. 
However, similar to previous work~\cite{evain21_interspeech,evain2021task}, we observe that training a feature extractor on more data~(3K and 7K hours) is beneficial for hybrid ASR, regardless of the pre-training data distribution~(see Table~\ref{tab:w2v2modelsdescription}). This relative low impact of biased pre-training data was also mentioned in Meng et al.~\cite{dontspeaktofast} for phoneme recognition. 
Lastly, we notice that the relative difference of WER between female and male talks ($\Delta_{rel}$) is not necessarily higher when gender-specific (male or female) pre-trained models are used ($\Delta_{rel}$ is 
 -12.3\% 
with the balanced pre-trained model 1K while it is 
 -8.2\%
with the male-only pre-trained model 1K).


\begin{table}
\centering
\caption{Hybrid~(a) and end-to-end~(b) ASR results~(WER) using the wav2vec~2.0 models either as feature extractors~(a) or speech encoders~(b). Results computed on the mTEDx test set. 
}
\scalebox{0.90}{
\begin{tabular}{l|c|c|c|c}
\toprule
\multicolumn{5}{c}{\textbf{(a) Hybrid ASR}} \\\midrule
\textbf{\multirow{2}{*}{Pre-training}}& \multicolumn{3}{c|}{\textbf{WER}}                  & \multicolumn{1}{c}{\multirow{2}{*}{$\mathbf{\Delta_{rel}},\%$}}   \\
\multicolumn{1}{c|}{\textbf{}}            & \textbf{M}     & \textbf{F}     & \textbf{All}   & \textbf{}      \\\midrule
F-1K-Large	&	25.7	&	22.3	&	24.3	&	-14.2	\\
M-1K-Large	&	25.4	&	23.4	&	24.8	&	-8.2	\\
LB-1K-Large	&	25.9	&	22.9	&	24.7	&	-12.3	\\
LB-3K-Large	&	\textbf{22.1}	&	\textbf{20.9}	&	\textbf{21.5}	&	-5.6	\\
LB-7K-Large	&	23.1	&	21.3	&	22.3	&	-8.1	
   \\\midrule
\multicolumn{5}{c}{\textbf{(b) End-to-end ASR}}  \\ \midrule
F-1K-Large & 20.9 & 17.7 & 19.5 & -16.9 \\
M-1K-Large & 21.0 & 18.5 & 19.9 & -12.7 \\
LB-1K-Large & \textbf{15.3} & \textbf{13.0} & \textbf{14.3} & -16.6  \\
LB-3K-Large & 15.5 & 13.5 & 14.6 & -13.9\\
LB-7K-Large & 15.9 & 13.2 & 14.7 & -19.0\\
\bottomrule
\end{tabular}}
\label{tab:resultsasr}
\end{table}

\subsection{End-to-end ASR}

Our five end-to-end ASR systems are implemented on the SpeechBrain toolkit~\cite{speechbrain}, being 
each composed of a wav2vec~2.0 module, a 1024-dimension dense hidden layer with a Leaky ReLU activation function, and a softmax output layer.
For each end-to-end model, 
the weights of the wav2vec~2.0 module were initialized from one of the pretrained models listed in Table~\ref{tab:w2v2modelsdescription}. The CTC loss function~\cite{graves2006connectionist} was used for 
training, and two different instances of Adam~\cite{adam} optimizers managed the weight updates: one dedicated to the wav2vec~2.0 module, the other one to the two additional layers. The output of the end-to-end model is based on characters: the vocabulary is composed of the $102$ lower-case symbols contained in the normalized manual transcriptions of the training set. The models were trained on the balanced dataset~(68\,h), and no LM was applied.

Results are presented in the bottom portion~(b) of Table~\ref{tab:resultsasr}. We observe that, different from the previous results~(a), the performance of the end-to-end ASR models seems to be very dependent on the balance of the dataset used to pre-train the SSL models. In these experiments, the model based on the wav2vec~2.0 with balanced pre-training data~(LB-1K-Large) resulted in the best results for both genders. Moreover, the models based on the gender-specific SSL models achieved poor performance, surprisingly even for the gender they targeted.\footnote{Due to a lack of space, we did not include all the 95\% confidence intervals. To give an idea of the statistical significance of these results, notice that for the first model, column \textsl{All}:
24.7\% WER $\in$ 
$[24.0,25.5]$ 
in (a), and
14.3\%
WER $\in$ 
$[13.2, 15.3]$
in (b).}
These results illustrate that, when fine-tuning an SSL model on the ASR task, the gender biases introduced during the pre-training are crucial for the downstream task, and cannot be fixed by including more data~(inferior performance of 3K and 7K). It also seems very important to consider the variability of speakers during the pre-training step: our results showed that the presence of speech for a given gender in the pretraining dataset helps to better transcribe speech for the opposite gender.

\section{Speech-to-Text Translation}\label{sec:st}

\begin{table*}
\caption{Speech translation performance (BLEU) for each pre-trained model and each language pair. Results obtained on the \textit{test} set of mTEDx. Scores in brackets show BLEU on separate [male, female] talks. $\Delta_{rel}=\frac{BLEU(female)-BLEU(male)}{0.5 \times (BLEU(female)+BLEU(male))}$}.
\label{tab:ST}
\centering
\scalebox{0.93}{
\begin{tabular}{l|cr|cr|cr}\toprule 
\textbf{Pre-training}                                      & \textbf{fr-en {[}M,F{]}}                  & \multicolumn{1}{r|}{$\mathbf{\Delta_{rel}},\%$}  & \textbf{fr-es {[}M,F{]}}                  & \multicolumn{1}{r|}{$\mathbf{\Delta_{rel}},\%$} & \textbf{fr-pt {[}M,F{]}}                  & \multicolumn{1}{r}{$\mathbf{\Delta_{rel}},\%$} \\
\midrule

F-1K-Large                                                  & 14.97 {[}14.34,15.71{]}          & +9.12               & 15.81 {[}15.71,15.99{]}          & +1.77             & 10.55 {[}12.00,8.56{]}           & -33.46             \\
M-1K-Large                                                      & 15.99 {[}15.90,16.11{]}          & +1.31                & 16.07 {[}15.55,16.75{]} & +7.43             & \textbf{12.01 {[}13.21,10.5{]}} & -22.86            \\
LB-1K-Large                                                  & 13.25 {[}12.62,14.09{]}          & +11.01               & 13.69 {[}13.37,14.08{]}          & +5.17             & 8.96 {[}9.73,7.93{]}             & -20.39            \\
LB-3K-Large                                                          & 17.44 {[}17.24,17.69{]} & +2.58              & 14.78 {[}14.84,14.63{]}       & -1.43             & 7.24 {[}8.07,6.12{]}             & -27.48             \\
LB-7K-Large                                                          & \textbf{17.50 {[}16.58,18.63{]}} & +11.64              & \textbf{16.34 {[}16.29,16.34{]}}       & +0.31             & 8.81 {[}9.83,7.42{]}             & -27.94             \\
\midrule
Table 5 of \cite{salesky21_interspeech} (bilingual e2e) & 8.9                              & \multicolumn{1}{l|}{-} & 10.6                             & \multicolumn{1}{l|}{-} & 7.9                              & \multicolumn{1}{l}{-}\\\bottomrule
\end{tabular}}
\end{table*}


We focus on direct speech-to-text translation, without producing any source language transcription. We use the SSL block as feature extractor. Our ST models follow Evain et al.~\cite{evain2021task}: we use the \textit{fairseq s2t} toolkit~\cite{wang2020fairseq} with their \texttt{s2t\_transformer\_xs} architecture~(Transformer~\cite{vaswani2017attention} with 6 encoder layers, 3 decoder layers, hidden dimension of 256). 
Following common practice~\cite{wang2020fairseq,inaguma2020espnet}, utterances with more than 3,000 frames are removed for GPU efficiency. All ST models are trained for 500 epochs using Adam~\cite{adam} and learning rate of \num{2e-3}.
We averaged last 10 checkpoints and used beam search (size of 5) decoding. Reported results are detokenized case-sensitive BLEU computed using sacreBLEU~\cite{post-2018-call} on test set. 
No specific ASR or MT pre-training (nor data augmentation) is applied as our goal is not to obtain best results, but to analyze impact of SSL pre-training. 
For extracting the speech features used as input of our ST models, we use all models from Table~\ref{tab:w2v2modelsdescription}.

Table~\ref{tab:ST} presents overall and separate BLEU on \textit{[male, female]} groups of TED talks, and the normalized relative difference of performance between female and male talks for all 15 ST models trained.\footnote{Note that since BLEU is used, the sign of the relative difference will be positive if female scores are better than male scores. This is the opposite to the calculation on
WER in Section~\ref{sec:asr}.} For reference, we also include the reported results from the original mTEDx paper~\cite{salesky21_interspeech}, which uses mel filterbank features instead of SSL, but also some data augmentation.
We observe that mTEDx dataset is challenging for direct ST~(low results for all three subsets). 
The fr-pt results are particularly low, variable and counter-intuitive: 3K and 7K models reach a lower performance compared to 1K, while the opposite is observed for fr-en and fr-es. The same trend difference was observed in previous work~\cite{evain2021task}, and we believe this might be sourced in the data scarcity for this language pair:~only 20\,h of speech available in the training set.

Focusing on models with the same amount of pre-training data~(1K), we observe medium variability of overall BLEU: for fr-en for instance, it ranges from 13.25~(balanced) to 15.99~(male),  depending on the SSL model used to extract features. Similar to the hybrid ASR experiments~((a) in Table~\ref{tab:resultsasr}) and previous work~\cite{dontspeaktofast}, we do not observe a gender-related performance issue in downstream models by using extremely unbalanced SSL models~(male and female) as
feature extractors in ST. Counterintuitively, the BLEU obtained with ST models that used features from these models is even better than the one obtained with the balanced model.
About the relative difference of BLEU between female and male talks~($\Delta_{rel}$), this metric is not higher when gender-specific SSL models are used: for fr-en, $\Delta_{rel}$ is +11.01\% with the balanced model, while it is only 1.31\% with the male-only model. This reinforces that the SSL feature extractors are not causing gender-related performance gap.
Moreover, we notice that the $\Delta_{rel}$ is very different from one language pair to another: it is positive for fr-en, and negative for fr-pt. This is particularly interesting considering that the test set is exactly the same, and only the target translation and the amount of training data differ. This suggests that there might exist other strong factors impacting ST performance, such as the target language, and gender distribution in the training sets.

\section{Discussion}\label{sec:discussion}

Our assessment of gender bias in SSL models was based on two different forms of downstream integration. 
When using our SSL blocks as simple feature extractors~(hybrid ASR and ST), we observe the same trend: results for gender-specific models were not worse than results with the balanced SSL model. 
This suggests that the wav2vec~2.0 features remain exploitable speech representations even if SSL models are trained on biased data. Further analysis is needed to understand the reasons behind this observation, but one possible explanation is that wav2vec~2.0 features contain less speaker-specific information. It was shown in Nguyen et al.~\cite{nguyen:hal-02962186} that speech representations obtained with contrastive predicting coding~(an ancestor of wav2vec~2.0) are less speaker-specific, and maybe this aspect is amplified by the quantization step that is part of the wav2vec~2.0 pipeline. 
Some more principled analysis such as the one of Pasad et al.~\cite{pasad2021layer}, which studies layer-wise representations from the wav2vec~2.0, would be needed to confirm this hypothesis. 
When the SSL block is used as a speech encoder in end-to-end ASR training, we find a different trend: using a well-balanced wav2vec~2.0 model leads to better overall 
performance. We also observe that all SSL models containing speakers from both genders in the pre-training data~(1K, 3K, 7K) achieve better results than the gender-specific models. 
This result illustrates that the interaction between pre-training and fine-tuning is complex. 
At this stage one can only formulate conjectures, but we hypothesize that a gender-balanced pre-training might provide a better initialization for the fine-tuning process, which itself relies on both male and female speech.

Regarding our basic `fairness' metric~(relative difference of performance measured between female and male test sets), it did not display strong variation from balanced to gender-specific pre-trained models. Many other factors may have more impact on performance such as language pair~(for ST), amount of training data for fine-tuning models~(ASR, ST), speech-to-text approach~(hybrid ASR versus end-to-end ASR), and even random seed used for model initialization (as shown in Garnerin et al.~\cite{garnerin-etal-2021-investigating}).\footnote{Due to the total number of models already trained for this study, the analysis of model stability using multiple runs was left for future work.}
We also find important to highlight a possible limitation in our investigation regarding speaker diversity in the French test set of mTEDx~(only 10 speakers). In future work we intend to extend our ASR experiments using a richer variability of speakers in the test set.
Finally, this investigation focused on wav2vec~2.0 architecture; our results are thus limited to a single SSL model and should be interpreted accordingly. 

Concluding, investigating gender bias in pre-training, fine-tuning, and inference for a speech-to-text pipeline is complex, and all these steps need to be carefully controlled. In this work we focused on the impact of the pre-training step. In the setting where a pre-trained model is used as a feature extractor, we observed the same trend for two downstream tasks~(hybrid ASR and ST): the impact of pre-training seems to be less important than other factors. However, in the setting where the pre-trained model is used to initialize a speech encoder, pre-training on a biased speech corpus may hurt the performance. This illustrates the non trivial interaction between pre-training and fine-tuning processes.
We believe that careful investigation of the layer-wise representations produced by these SSL models might help us better understand these aspects.

\noindent
\section{Acknowledgements}
This work used HPC resources from GENCI-IDRIS (2020-A0111012991, 2021-AD011013317 and 2021-AD011013331). It was also 
funded by the European Commission through the SELMA project under grant number 957017.

\bibliographystyle{IEEEtran}

\bibliography{shortened_bib} 

\end{document}